%% file: main.tex
\documentclass[10pt,twocolumn,letterpaper]{article}

 \usepackage{cvpr}              


\definecolor{cvprblue}{rgb}{0.21,0.49,0.74}
\usepackage[pagebackref,breaklinks,colorlinks,allcolors=cvprblue]{hyperref}

\usepackage{pgfplots}
\pgfplotsset{compat=1.18}

\usepackage{pgfplotstable}
\usepgfplotslibrary{groupplots}

\usepackage{booktabs}
\usepackage{multirow}
\usepackage{graphicx}
\usepackage{tikz}
\usetikzlibrary{positioning,fit,calc,arrows.meta}

\usetikzlibrary{backgrounds}

\definecolor{cBlue}{RGB}{55,126,184}
\definecolor{cLightOrange}{RGB}{253,174,107}
\definecolor{cOrange}{RGB}{230,85,13}

\def\paperID{9}
\def\confName{CVPR}
\def\confYear{2026}

\title{Token-Space Mask Prediction for Efficient Vision Transformer Segmentation}

\author{
Calvin Galagain$^{1,2}$ \quad Martyna Poreba$^{1}$ \quad François Goulette$^{2}$\\
$^{1}$Université Paris-Saclay, CEA List \quad
$^{2}$U2IS, ENSTA Paris, Institut Polytechnique de Paris\\
F-91120 Palaiseau, France\\
{\tt\small \{calvin.galagain, martyna.poreba\}@cea.fr \quad francois.goulette@ensta.fr}
}

\begin{document}
\maketitle
\input{content/0_abstract}
\input{content/1_intro_mp}
\input{content/2_related_mp}

\input{content/3_method_mp}
\input{content/4_experiments}

\input{content/5_ablations}
\input{content/6_conclusions}

\section*{Acknowledgments}
This work was partially funded by the French Defence Innovation
Agency (Agence de l'innovation de d\'efense, AID).

{
    \small
    \bibliographystyle{ieeenat_fullname}
    \bibliography{main}
}


\end{document}

%% file: content/0_abstract.tex
\begin{abstract}
Query-based Vision Transformer segmentation models typically reconstruct dense spatial feature maps to predict masks, inheriting design patterns from convolutional architectures. We show that this explicit image-space reconstruction is not required. We introduce TokenMask, a token-space mask head that computes mask logits directly from query-token affinities and performs interpolation in logit space rather than feature space. This reformulation preserves the original linear scoring mechanism while simplifying the computational structure. Across diverse ViT backbones, datasets and segmentation tasks, TokenMask consistently improves efficiency over prior approaches by reducing computational and memory requirements while maintaining competitive accuracy, leading to tangible speedups on NVIDIA Jetson AGX Orin using TensorRT FP16 inference.
Overall, TokenMask yields a simpler and more deployment-friendly design for embedded vision systems.
\end{abstract}

%% file: content/1_intro_mp.tex
\section{Introduction}
\label{sec:intro}
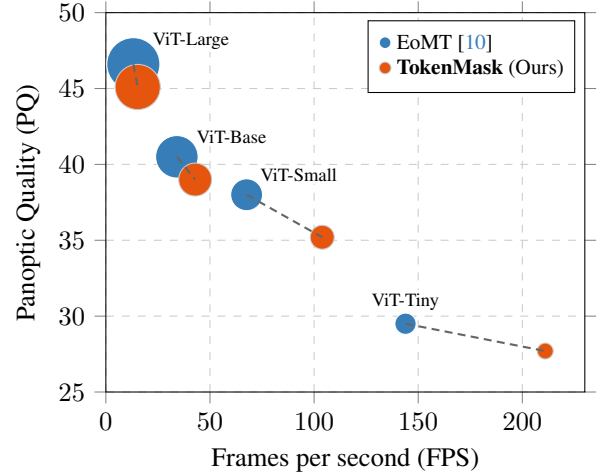
\begin{figure}[t]
\centering
\begin{tikzpicture}
\begin{axis}[
    width=0.95\linewidth,
    height=6.6cm,
    xmin=0, xmax=230,
    ymin=25, ymax=50,
    xlabel={Frames per second (FPS)},
    ylabel={Panoptic Quality (PQ)},
    grid=both,
    major grid style={dashed, gray!40},
    minor grid style={dotted, gray!25},
    tick align=outside,
    clip=false,
    set layers=standard,
]


\addplot[
    only marks,
    mark=*,
    mark size=4.0pt,
    mark options={draw=white, fill=cBlue}
] coordinates {(143.9,29.5)};

\addplot[
    only marks,
    mark=*,
    mark size=6.0pt,
    mark options={draw=white, fill=cBlue}
] coordinates {(67.6,38.0)};

\addplot[
    only marks,
    mark=*,
    mark size=8.0pt,
    mark options={draw=white, fill=cBlue}
] coordinates {(34.1,40.5)};

\addplot[
    only marks,
    mark=*,
    mark size=10.0pt,
    mark options={draw=white, fill=cBlue}
] coordinates {(13.2,46.6)};

\addplot[
    only marks,
    mark=*,
    mark size=3.0pt,
    mark options={draw=black!20, fill=cOrange}
] coordinates {(211.0,27.7)};

\addplot[
    only marks,
    mark=*,
    mark size=4.5pt,
    mark options={draw=black!20, fill=cOrange}
] coordinates {(103.9,35.2)};

\addplot[
    only marks,
    mark=*,
    mark size=6.2pt,
    mark options={draw=black!20, fill=cOrange}
] coordinates {(42.9,39.0)};

\addplot[
    only marks,
    mark=*,
    mark size=8.5pt,
    mark options={draw=black!20, fill=cOrange}
] coordinates {(15.3,45.1)};

\addplot[
    no marks,
    densely dashed,
    thick,
    black!60,
    on layer=axis foreground
] coordinates {(143.9,29.5) (211.0,27.7)};

\addplot[
    no marks,
    densely dashed,
    thick,
    black!60,
    on layer=axis foreground
] coordinates {(67.6,38.0) (103.9,35.2)};

\addplot[
    no marks,
    densely dashed,
    thick,
    black!60,
    on layer=axis foreground
] coordinates {(34.1,40.5) (42.9,39.0)};

\addplot[
    no marks,
    densely dashed,
    thick,
    black!60,
    on layer=axis foreground
] coordinates {(13.2,46.6) (15.3,45.1)};

\node[font=\scriptsize, fill=white, inner sep=1pt, anchor=south]
    at (axis cs:143.9,30.2) {ViT-Tiny};

\node[font=\scriptsize, fill=white, inner sep=1pt, anchor=south west, xshift=4pt]
    at (axis cs:67.6,38.7) {ViT-Small};

\node[font=\scriptsize, fill=white, inner sep=1pt, anchor=south west, xshift=5pt]
    at (axis cs:35.8,41.2) {ViT-Base};

\node[font=\scriptsize, fill=white, inner sep=1pt, anchor=south west, xshift=5pt]
    at (axis cs:14.6,47.4) {ViT-Large};

\node[
    draw=black,
    fill=white,
    font=\footnotesize,
    anchor=north east,
    inner sep=4pt
] at (rel axis cs:0.98,0.98) {%
\begin{tabular}{@{}l@{}}
\tikz{\filldraw[draw=white,fill=cBlue] (0,0) circle (2.2pt);}~EoMT~\cite{Kerssies2025YourVI} \\
\tikz{\filldraw[draw=black!20,fill=cOrange] (0,0) circle (2.2pt);}~\textbf{TokenMask} (Ours)
\end{tabular}
};

\end{axis}
\end{tikzpicture}
\caption{
Efficiency-accuracy trade-off on ADE20K panoptic segmentation.
Marker size is proportional to the number of GFLOPs. Dashed connectors show the transition from EoMT to TokenMask for each backbone size.
}
\label{fig:teaser}
\end{figure}

Visual segmentation is a core problem in computer vision, aiming to assign semantic labels to every pixel in an image. It is typically studied under three settings: semantic, instance, and panoptic segmentation. These differ in their level of granularity, with panoptic segmentation unifying the former two by jointly modeling \textit{stuff} and \textit{thing} categories.
These capabilities are critical for embedded vision applications such as robotics and intelligent systems, where accurate scene understanding must be achieved under strict real-time and computational constraints~\cite{galagain2025semanticslamreadyembedded}.

Recent advances in deep learning have significantly improved segmentation performance, notably through the introduction of Vision Transformers (ViTs)~\cite{dosovitskiy2020image}. In this context, query-based approaches have emerged as a powerful paradigm~\cite{Carion2020EndtoEndOD}, representing objects or regions as a set of learnable queries and enabling unified, global reasoning across different segmentation tasks. However, despite their conceptual appeal, existing query-based segmentation architectures remain computationally heavy and structurally complex, limiting their applicability to embedded platforms. Many existing query-based segmentation models rely on multi-stage decoding pipelines that combine dense pixel decoders with multi-layer transformer query decoders~\cite{Cheng2021PerPixelCI, Cheng2021MaskedattentionMT}. While effective, this design couples token-based representations with explicit spatial decoding stages, introducing additional complexity that is not intrinsic to ViT-based segmentation. In particular, although these models operate on token sequences internally, mask prediction is still commonly performed in image space. Patch tokens are first reconstructed into dense spatial feature maps, often followed by feature upsampling to recover spatial resolution, before computing spatial interactions with segmentation queries. 
This design mirrors convolutional segmentation pipelines and reflects architectural conventions rather than a fundamental modeling requirement, leading to increased memory footprint and latency that are problematic for real-time embedded deployment.

This observation motivates a rethinking of mask prediction in query-based segmentation models. We move away from explicit spatial reconstruction and instead align the segmentation head with the token-based representation inherent to ViTs. From this perspective, we revisit the role of upsampling under resource constraints. We introduce TokenMask, a unified decoding head that can be seamlessly attached to a ViT backbone, and jointly supports instance, panoptic and semantic segmentation. As illustrated in Fig.~\ref{fig:teaser}, TokenMask preserves segmentation quality across ViT backbones while reducing computation by up to ~50–55\% compared to EoMT~\cite{Kerssies2025YourVI}. It also consistently improves throughput, with up to ~45–50\% higher FPS on Nvidia Jeston Orin. 

Our contributions are threefold: 
\textbf{(1)} We reformulate mask prediction for ViT-based segmentation directly in token space, eliminating explicit spatial feature-map reconstruction while preserving the original linear scoring mechanism;
\textbf{(2)} We analyze the impact of relocating upsampling from feature space to logit space and investigate regimes in which upsampling can be removed entirely; 
\textbf{(3)} We conduct a systematic evaluation across multiple backbones, datasets, and output strides, highlighting the computational and memory implications of mask-head design choices for embedded deployment.

%% file: content/2_related_mp.tex
\section{Related Work}
\label{sec:related}

Semantic segmentation with ViTs has evolved primarily through the design of segmentation heads. Early approaches such as SETR~\cite{SETR,Zhang_2024} rely on CNN-based upsampling heads to recover spatial predictions from patch tokens. In contrast, Segmenter~\cite{strudel2021segmenter} and StructToken~\cite{StructToken} introduce fully Transformer-based heads, leveraging learnable class or structure tokens that interact with patch embeddings to directly produce segmentation masks. SegFormer~\cite{xie2021segformersimpleefficientdesign}  adopts a different strategy with a lightweight MLP head that efficiently fuses multi-scale features. More recently, SegViT \cite{zhang2022segvitsemanticsegmentationplain} and SegViT2 \cite{zhang2023segvitv2exploringefficientcontinual} depart from explicit decoding by proposing an attention-to-mask mechanism, where class tokens directly generate segmentation masks from similarity maps.

Compared to dense ViT-based segmentation, query-based approaches predict a structured set of object and region masks rather than per-pixel labels. This formulation reduces heuristic post-processing, enables multi-task unification, and provides explicit control over complexity via the number of queries. It also promotes object-level reasoning and more efficient computation by focusing on a limited set of relevant regions instead of dense token-wise predictions. The first query-based segmentation approaches build upon DETR~\cite{Carion2020EndtoEndOD} by extending its set-based prediction framework with mask prediction heads, allowing each learned query to generate an object mask through interactions with dense image features. 
In these approaches, mask prediction is typically formulated in image space by projecting query embeddings onto spatial feature maps, tightly coupling query-based reasoning with dense spatial representations.

Building on this paradigm, MaskFormer~\cite{Cheng2021PerPixelCI} reformulates semantic segmentation as a mask classification problem, predicting a set of class-labeled masks using learnable queries. Mask2Former~\cite{Cheng2021MaskedattentionMT} extends query-based segmentation to a unified setting covering semantic, instance, and panoptic tasks by introducing a dense pixel decoder coupled with a multi-layer transformer query decoder for progressive query refinement. Mask prediction is achieved by associating each query with a learned mask representation that interacts with dense spatial features to produce segmentation outputs. Although this interaction is implemented in image space via dense feature maps, it effectively reduces to a linear similarity operation between query representations and spatially organized features. Mask2Former has become a reference architecture for query-based segmentation, and several subsequent works build upon it while targeting different objectives. OneFormer~\cite{Jain2022OneFormerOT} extends the framework with task-conditioned queries to enable unified multi-task segmentation without architectural changes. EoMT~\cite{Kerssies2025YourVI} targets efficiency by simplifying the query-based segmentation pipeline. It removes several task-specific components commonly used in Mask2Former architecture, including the pixel decoder and the transformer query decoder, and repurposes the ViT encoder itself to jointly process image tokens and object queries. EoMT enables one-stage segmentation with significantly lower latency, while still relying on image-space mask prediction. 
Recently, LiPS~\cite{galagain2026lipslightweightpanopticsegmentation} identifies a structural inefficiency in Mask2Former-like architectures, where the majority of computation is devoted to dense multi-scale feature processing rather than query-based reasoning. To address this, it preserves the masked transformer decoding paradigm while simplifying the upstream pipeline through selective feature routing and spatial compression, leading to a substantially improved accuracy-efficiency trade-off.

%% file: content/3_method_mp.tex
\section{Methodology}
\label{sec:method}

Existing query-based segmentation architectures ~\cite{Kerssies2025YourVI, Cheng2021MaskedattentionMT}, retain image-space mask heads inherited from dense decoders, requiring spatial reconstruction and feature upsampling. This design increases memory footprint and limits scalability. To address this, we propose TokenMask, a lightweight token-space segmentation head for embedded ViTs that reduces memory and computational overhead while supporting unified semantic, instance, and panoptic segmentation.



\begin{figure*}[t]
\centering
\resizebox{\linewidth}{!}{
\begin{tikzpicture}[
  font=\small,
  node distance=6mm and 10mm,
  box/.style={draw, rounded corners=2pt, align=center, inner sep=5pt, minimum width=30mm},
  op/.style={draw, rounded corners=2pt, align=center, inner sep=5pt, minimum width=34mm, fill=gray!8},
  important/.style={draw, rounded corners=2pt, align=center, inner sep=5pt, minimum width=34mm, fill=red!8},
  good/.style={draw, rounded corners=2pt, align=center, inner sep=5pt, minimum width=34mm, fill=green!10},
  arrow/.style={-Latex, thick},
  frameRed/.style     ={draw=red, dashed, line width=0.9pt, rounded corners=2pt},
  frameBlue/.style    ={draw=blue, dashed, line width=0.9pt, rounded corners=2pt},
  frameGreen/.style   ={draw=green!70!black, dashed, line width=0.9pt, rounded corners=2pt},
  frameOrange/.style  ={draw=orange!90!black, dashed, line width=0.9pt, rounded corners=2pt},
  frameMagenta/.style ={draw=magenta, dashed, line width=0.9pt, rounded corners=2pt},
  tag/.style={font=\bfseries\large}
]

\begin{scope}[on background layer]


  \coordinate (A1) at (-20mm,  5mm); 
  \coordinate (A2) at (20mm,  5mm); 
  \coordinate (A3) at (20mm, -32mm); 
  \coordinate (A4) at (-20mm, -32mm); 
  \draw[frameRed] (A1)--(A2)--(A3)--(A4)--cycle;

  \coordinate (B1) at (-20mm, -34mm); 
  \coordinate (B2) at (20mm, -34mm); 
  \coordinate (B3) at (20mm, -80mm); 
  \coordinate (B4) at (-20mm, -80mm); 
  \draw[frameBlue] (B1)--(B2)--(B3)--(B4)--cycle;

  \coordinate (C1) at ( 27mm,-19mm);
  \coordinate (C2) at ( 67mm,-19mm);
  \coordinate (C3) at ( 67mm,-97mm);
  \coordinate (C4) at (-20mm,-97mm);
  \coordinate (C5) at (-20mm,-83mm);
  \coordinate (C6) at ( 27mm,-83mm);
  \draw[frameGreen] (C1)--(C2)--(C3)--(C4)--(C5)--(C6)--cycle;

  \def\xRL{80mm}   
  \def\xRR{170mm}  

  \coordinate (AR1) at (\xRL,  9mm);    
  \coordinate (AR2) at (125mm,  9mm);    
  \coordinate (AR3) at (125mm, -28mm);   
  \coordinate (AR4) at (\xRL, -28mm);   
  \draw[frameRed] (AR1)--(AR2)--(AR3)--(AR4)--cycle;

  \coordinate (BR1) at (130mm, -15mm);
  \coordinate (BR2) at (\xRR, -15mm);
  \coordinate (BR3) at (\xRR, -48mm);
  \coordinate (BR4) at (\xRL, -48mm);
  \coordinate (BR5) at (\xRL, -30mm);
  \coordinate (BR6) at (130mm, -30mm);
  \draw[frameOrange] (BR1)--(BR2)--(BR3)--(BR4)--(BR5)--(BR6)--cycle;

  \coordinate (CR1) at (\xRL, -50mm);
  \coordinate (CR2) at (125mm, -50mm);
  \coordinate (CR3) at (125mm,-96mm);
  \coordinate (CR4) at (\xRL,-96mm);
  \draw[frameMagenta] (CR1)--(CR2)--(CR3)--(CR4)--cycle;

\end{scope}

\def\panelgap{18mm}

\node[box] (imgA) {Image};
\node[op, below=of imgA] (vitA) {ViT backbone};
\node[box, below=of vitA] (tokensA) {Patch tokens $T$\\$\mathbb{R}^{B\times N\times C}$};
\node[op, below=of tokensA] (reshapeA) {Token-to-Grid \\ reshape};
\node[box, below=of reshapeA] (featA) {Feature map $F$\\$\mathbb{R}^{B\times C\times H\times W}$};
\node[important, below=of featA] (upFeatA) {Feature upsample\\$C$ channels};
\node[op, below=of upFeatA] (dotA) {Spatial dot-product\\$M=\langle m(Q),F\rangle$};
\node[box, below=of dotA] (outA) {Mask logits\\$\mathbb{R}^{B\times Q\times \hat H\times \hat W}$};

\node[box, right=18mm of tokensA] (qryA) {Queries $Q$\\$\mathbb{R}^{B\times Q\times C}$};
\node[op, below=of qryA] (mheadA) {MaskHead $m(\cdot)$\\$\mathbb{R}^{B\times Q\times C}$};

\draw[arrow] (imgA) -- (vitA);
\draw[arrow] (vitA) -- (tokensA);
\draw[arrow] (tokensA) -- (reshapeA);
\draw[arrow] (reshapeA) -- (featA);
\draw[arrow] (featA) -- (upFeatA);
\draw[arrow] (upFeatA) -- (dotA);
\draw[arrow] (dotA) -- (outA);

\draw[arrow] (qryA) -- (mheadA);
\draw[arrow] (mheadA.south) |- (dotA.east);

\node[fit=(imgA)(outA)(qryA)(mheadA), inner sep=4mm] (frameA) {};

\node[box, anchor=north west] (imgB) at ($(frameA.north east)+(\panelgap,0)$) {Image};
\node[op, below=of imgB] (vitB) {ViT backbone};
\node[box, below=of vitB] (tokensB) {Patch tokens $T$\\$\mathbb{R}^{B\times N\times C}$};

\node[box, right=18mm of tokensB] (qryB) {Queries $Q$\\$\mathbb{R}^{B\times Q\times C}$};
\node[op, below=of qryB, yshift=-2mm] (mheadB) {MaskHead $m(\cdot)$\\$\mathbb{R}^{B\times Q\times C}$};

\node[good, below=of tokensB] (matmulB) {Token-space scores\\$L=m(Q)\,T^\top$\\$\mathbb{R}^{B\times Q\times N}$};
\node[op, below=of matmulB] (reshapeB) {Token-score-to-Grid reshape\\$N \rightarrow H\times W$};
\node[box, below=of reshapeB] (maskPatchB) {Mask logits (token grid)\\$\mathbb{R}^{B\times Q\times H\times W}$};

\node[good, below=of maskPatchB] (upLogitB) {Logit upsample\\$Q$ channels};
\node[box, below=of upLogitB] (outB) {Mask logits\\$\mathbb{R}^{B\times Q\times \hat H\times \hat W}$};

\draw[arrow] (imgB) -- (vitB);
\draw[arrow] (vitB) -- (tokensB);

\draw[arrow] (qryB) -- (mheadB);
\draw[arrow] (tokensB) -- (matmulB);
\draw[arrow] (mheadB.west) |- (matmulB.east);

\draw[arrow] (matmulB) -- (reshapeB);
\draw[arrow] (reshapeB) -- (maskPatchB);
\draw[arrow] (maskPatchB) -- (upLogitB);
\draw[arrow] (upLogitB) -- (outB);

\node[fit=(imgB)(outB)(qryB)(mheadB), inner sep=6mm] (frameB) {};

\coordinate (titleLine) at ($(frameA.north)!0.5!(frameB.north)$);

\node[font=\bfseries, anchor=south]
  at ($(frameA.north |- titleLine)+(0,2mm)$)
  {(a) Mask head (image-space)};

\node[font=\bfseries, anchor=south]
  at ($(frameB.north |- titleLine)+(0,2mm)$)
  {(b) Proposed TokenMask head (token-space)};

\node[font=\small, anchor=north, align=center]
  at ($(frameA.south)+(0,-2mm)$)
  {reshape $\rightarrow$ feature map $\rightarrow$ spatial ops};

\node[font=\small, anchor=north, align=center]
  at ($(frameB.south)+(0,-2mm)$)
  {no feature map; mask via token scores};

\node[tag, text=red,   anchor=west]    at ($(A2)+(1mm,0mm)$) {A};
\node[tag, text=blue,  anchor=west]    at ($(B2)!0.25!(B3)$) {B};
\node[tag, text=green!70!black, anchor=north] at ($(C3)!0.25!(C4)$) {C};

\node[tag, text=red,  anchor=west] at ($(AR2)+(1mm,0mm)$) {A};
\node[tag, text=orange!90!black, anchor=west] at ($(BR2)!0.25!(BR3)$) {D};
\node[tag, text=magenta, anchor=west] at ($(CR2)!0.5!(CR3)$) {E};

\draw[arrow] (tokensA) -- (qryA) node[midway, above, sloped] {\scriptsize inject / concat};
\draw[arrow] (tokensB) -- (qryB) node[midway, above, sloped] {\scriptsize inject / concat};

\end{tikzpicture}
}
\caption{
    ViT backbone interfacing with Mask (image-space) and proposed TokenMask (token-space mask) heads:
    (a) Classical design reshapes patch tokens into an image-space feature map and computes masks via a dense spatial dot-product after upsampling $C$-channel features. 
    (b) Proposed TokenMask head. Mask logits are computed directly in token space via $L = m(Q)T^\top$ and upsampled in logit space over $Q$ channels, eliminating the need for explicit image-space feature maps. 
    \textbf{A} corresponds to tokenization and ViT backbone processing that produce patch tokens. 
    \textbf{B} denotes spatial reconstruction for mask prediction through token-to-grid reshaping and feature-space upsampling. 
    \textbf{C} indicates query-based mask prediction via spatial dot-product in image space. 
    \textbf{D} represents direct token-level scoring through query-token affinity computation. 
    \textbf{E} corresponds to reshaping token-level scores to the spatial grid followed by logit-space upscaling.
    }
\label{fig:architecture}
\end{figure*}
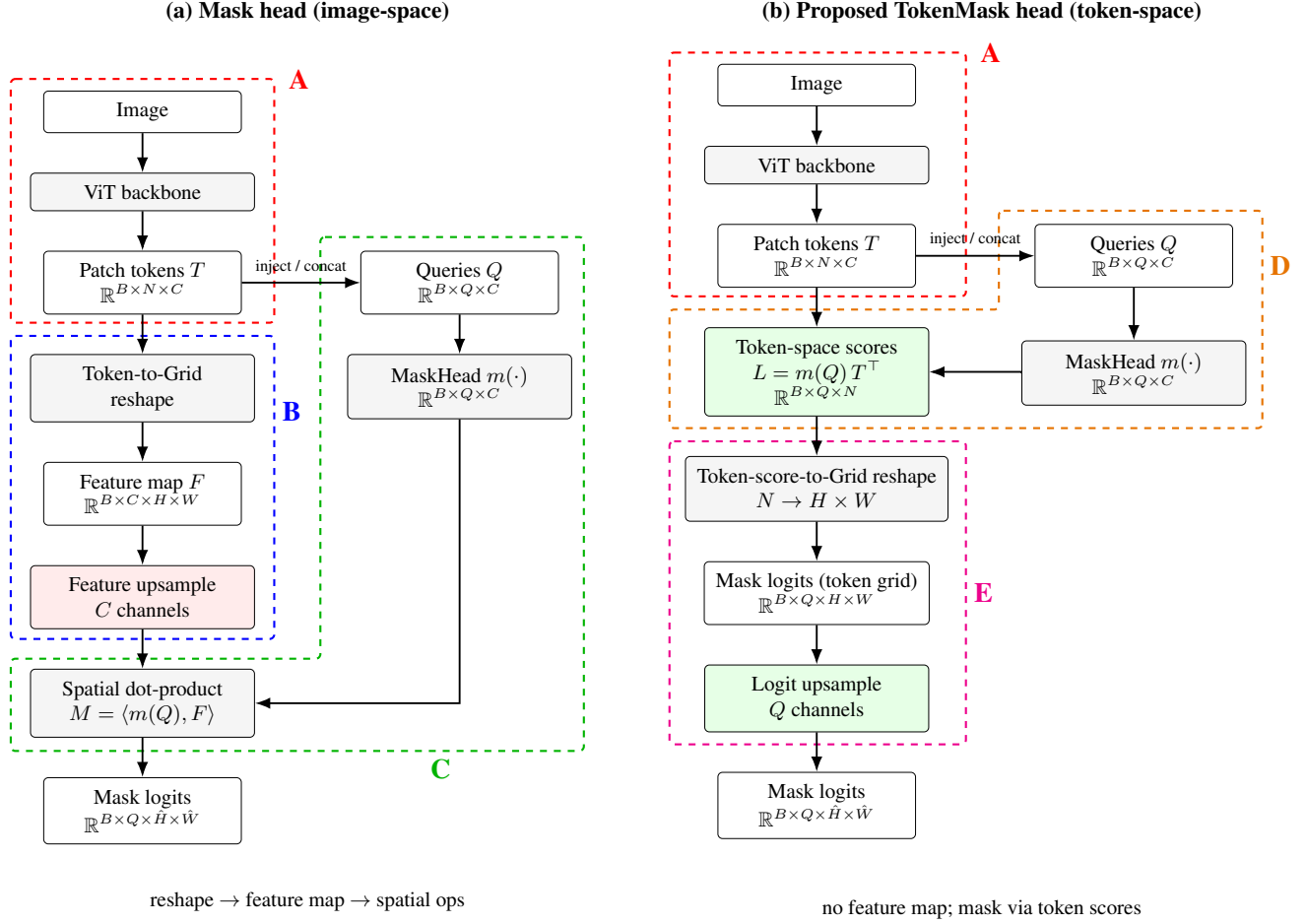


\subsection{Image-Space Mask Head}

\paragraph{Tokenization and ViT backbone.}
Let the input image be
$x \in \mathbb{R}^{B \times 3 \times H_0 \times W_0}$
where $B$ denotes the batch size and $H_0, W_0$ the input spatial resolution. The image is partitioned into non-overlapping patches that are linearly projected into $C$-dimensional embeddings and processed by a ViT backbone. The backbone outputs a sequence of patch tokens:
\begin{equation}
T \in \mathbb{R}^{B \times N \times C},
\end{equation}
where $N = H_p \times W_p$ denotes the patch-grid resolution. Each token encodes visual content from a specific spatial region together with positional information, forming the core internal representation used by ViT-based segmentation models.

\paragraph{Spatial reconstruction for mask prediction.}

In most query-based segmentation architectures, mask prediction is performed in image space.
As illustrated in Fig.~\ref{fig:architecture} (a), the patch tokens $T$ are first reshaped or rearranged into a dense spatial feature map:
\begin{equation}
F = \mathrm{reshape}(T) \in \mathbb{R}^{B \times C \times H_p \times W_p}.
\end{equation}
This reshape operation introduces no new information and merely restores an image-like tensor organization from the token sequence. The resulting feature map $F$ is typically processed by a pixel decoder and optionally upsampled along the spatial dimensions to recover resolution:
\begin{equation}
\tilde{F} \in \mathbb{R}^{B \times C \times \hat{H} \times \hat{W}},
\end{equation}
where $\hat{H}$ and $\hat{W}$ denote the upsampled spatial dimensions. Upsampling is applied per-channel; hence compute and memory scale linearly with the channel dimension $C$.

\paragraph{Query-based mask prediction.}

Query-based models maintain a fixed set of learnable queries:
\begin{equation}
Q \in \mathbb{R}^{B \times Q \times C},
\end{equation}
where $Q$ denotes the number of queries.
The queries are injected into a transformer decoder and refined through interactions with image features, yielding query representations used for mask prediction. Each query is projected by a mask head into a mask embedding, and segmentation masks are obtained via a spatial dot-product with the feature map $F$:
\begin{equation}
M = m(Q) \cdot F,
\end{equation}
where $m(\cdot)$ denotes a learned linear projection.
The resulting tensor:
\begin{equation}
M \in \mathbb{R}^{B \times Q \times H_p \times W_p}
\end{equation}
corresponds to the mask logits for all queries, which are subsequently used for loss computation and final segmentation decoding.
Although implemented as a spatial operation over dense feature maps, this mechanism fundamentally corresponds to a linear interaction between query embeddings and spatially organized token representations.

\subsection{Token-Space Mask Head: TokenMask}

\paragraph{Direct token-level scoring.}

As illustrated in Fig.~\ref{fig:architecture} (b), we reformulate mask prediction directly in token space and remove the explicit reconstruction of spatial feature maps. Given the patch tokens extracted by the ViT backbone and the projected query embeddings from the mask head, we compute mask scores directly at the token level via a matrix multiplication:
\begin{equation}
L = m(Q) T^\top,
\end{equation}
yielding:
\begin{equation}
L \in \mathbb{R}^{B \times Q \times N}.
\end{equation}
Each entry:
\begin{equation}
L_{q,i} = \langle m(q), t_i \rangle
\end{equation}
represents the affinity between query $q$ and patch token $i$, and can be interpreted as a membership score indicating how strongly token $i$ contributes to the mask predicted by query $q$. This operation is strictly equivalent to the spatial dot-product used in image-space mask heads  (Fig.~\ref{fig:architecture} (a)), but is performed directly on the token sequence without intermediate spatial reconstruction.

\paragraph{Reshaping and logit-space upsampling.}

The token-level scores are then reshaped back to the patch grid:
\begin{equation}
M_{\text{patch}} \in \mathbb{R}^{B \times Q \times H_p \times W_p},
\end{equation}
recovering a coarse spatial layout aligned with the patch resolution. To obtain the final mask predictions, interpolation is applied directly in logit space:
\begin{equation}
M \in \mathbb{R}^{B \times Q \times \hat{H} \times \hat{W}}.
\end{equation}

Importantly, the proposed formulation does not modify the underlying linear scoring mechanism. 
The mask remains a projection between query embeddings and local representations.

\subsection{Implications for Embedded Deployment}
By design, TokenMask eliminates feature-space upsampling and operates exclusively on compact token{-}level representations. This choice substantially reduces memory access cost and intermediate activation storage, two dominant bottlenecks in embedded vision pipelines where off{-}chip memory access often outweighs arithmetic cost.

First, TokenMask avoids explicit spatial feature reconstruction, thereby preventing the formation of large intermediate feature maps whose size scales with both spatial resolution and channel dimensionality $C$, which can be large in practice (e.g., $C \in [384, 1024]$). In conventional pipelines, this reconstruction step reorganizes token sequences into dense, image{-}like tensors that must be materialized in memory, leading to increased memory footprint and substantial data movement. In contrast, TokenMask computes mask scores directly in token space through a matrix multiplication between projected queries and patch tokens. This computation maps naturally to batched general matrix{-}matrix multiplication (GEMM), a highly optimized primitive on modern hardware accelerators, including embedded GPUs, NPUs, and DSPs. Leveraging GEMM enables efficient use of vector units, high arithmetic intensity, and favorable data reuse within on{-}chip memory, leading to improved throughput and reduced energy per inference.

Moreover, relocating upsampling from feature space to logit space further improves efficiency, as interpolation is performed over $Q$ channels rather than $C$, with typically $Q \ll C$. This decouples the computational cost of mask prediction from the backbone embedding dimension and yields a segmentation head whose complexity scales more favorably with model capacity and input resolution. 

%% file: content/4_experiments.tex

\begin{table*}[t]
\centering
\footnotesize
\setlength{\tabcolsep}{6pt}
\renewcommand{\arraystretch}{1.15}

\caption{Performance and efficiency comparison on ADE20K for panoptic and semantic segmentation.
Panoptic results are reported at $640\times640$ input resolution, and semantic results at $512\times512$.}
\label{tab:panoptic_semantic}

\begin{tabular}{l l cccc cccc}
\toprule
\multicolumn{2}{c}{} 
& \multicolumn{4}{c}{\textbf{Panoptic}} 
& \multicolumn{4}{c}{\textbf{Semantic}} \\
\cmidrule(lr){3-6} \cmidrule(lr){7-10}

\textbf{Backbone} & \textbf{Method}
& \textbf{FPS A100} $\uparrow$
& \textbf{FPS Orin} $\uparrow$
& \textbf{GFLOPs} $\downarrow$
& \textbf{PQ} $\uparrow$
& \textbf{FPS A100} $\uparrow$
& \textbf{FPS Orin} $\uparrow$
& \textbf{GFLOPs} $\downarrow$
& \textbf{mIoU} $\uparrow$ \\

\midrule

ResNet-50
& Mask2Former
& 0.6 & 7.1 & 147.2 & 39.6
& 0.6 & 7.1 & 147.2 & 46.0 \\

\midrule

AFFormer~\cite{Dong2023HeadFreeLS}
& LiPS
& 3.6 & 17.5 & 26.4 & 35.8
& 3.6 & 17.5 & 26.4 & 44.0 \\

\midrule

ViT-Tiny
& Segmenter
& - & - & - & -
& 147.4 & 69.0 & 12.8 & 40.0 \\
& EoMT
& 189.0 & 143.9 & 20.8 & 29.5
& 192.4 & 248.0 & 13.6 & 40.0 \\
& \textbf{TokenMask} (Ours)
& 202.9 & 211.0 & 9.7 & 27.7
& 211.0 & 363.6 & 6.4 & 39.1 \\

\midrule

ViT-Small
& Segmenter
& - & - & - & -
& 94.9 & 69.0 & 38.6 & 46.5 \\
& EoMT
& 129.3 & 67.6 & 71.7 & 38.0
& 133.0 & 114.4 & 46.7 & 45.2 \\
& \textbf{TokenMask} (Ours)
& 150.8 & 103.9 & 37.3 & 35.2
& 151.8 & 175.4 & 24.6 & 43.3 \\

\midrule

ViT-Base
& Segmenter
& - & - & - & -
& 43.2 & 43.4 & 129.6 & 49.6 \\
& EoMT
& 94.8 & 34.1 & 263.2 & 40.5
& 118.4 & 57.6 & 171.8 & 46.5 \\
& \textbf{TokenMask} (Ours)
& 131.6 & 42.9 & 146.3 & 39.0
& 150.3 & 75.6 & 96.53 & 46.9 \\

\midrule

ViT-Large
& Segmenter
& - & - & - & -
& 15.0 & 18.9 & 400.1 & 52.3 \\
& EoMT
& 48.5 & 13.2 & 699.9 & 46.6
& 64.6 & 21.7 & 454.2 & 54.8 \\
& \textbf{TokenMask} (Ours)
& 61.9 & 15.3 & 501.5 & 45.1
& 81.5 & 24.4 & 326.3 & 53.1 \\

\bottomrule
\end{tabular}
\end{table*}

\section{Experiments}

We evaluate the proposed TokenMask head by analyzing its impact on both segmentation accuracy and computational efficiency. Unless stated otherwise, all models share identical initialization, query count, decoder depth, and training protocol; observed differences therefore stem from the mask prediction design and where upsampling is applied. Our experimental analysis follows two axes. First, in the main results we compare the image-space mask head used in EoMT with our token-space formulation. 
Second, in ablation studies we isolate the role of upsampling by comparing feature-space upsampling, logit-space upsampling applied after mask scoring, and a no-upsampling configuration where predictions are produced at native patch-grid resolution.

\subsection{Experimental Setup}

\paragraph{Datasets.} 


We evaluate our approach on three widely used panoptic segmentation benchmarks spanning diverse scene types and annotation granularities. ADE20K~\cite{zhou2017scene} contains 150 semantic categories across indoor and outdoor environments, and we use its official panoptic training and validation splits. Cityscapes~\cite{Cordts2016TheCD} focuses on high-resolution urban street scenes with fine-grained semantic and instance annotations; results are reported on the standard validation set. COCO-Panoptic~\cite{lin2014microsoft} combines instance-level and semantic annotations over complex natural scenes and serves as a large-scale benchmark for unified panoptic segmentation.


\paragraph{Evaluation Metrics.}
Computational efficiency is assessed using both theoretical complexity (GFLOPs) and empirical runtime measured in frames per second (FPS). To ensure fair and reproducible runtime evaluation across platforms, we report FPS with batch size one on an NVIDIA A100~\footnote{\url{https://www.nvidia.com/en-us/data-center/a100/}} using PyTorch FP32 inference as a standardized high-performance reference, and the target embedded platform, an NVIDIA Jetson AGX Orin~\footnote{\url{https://developer.nvidia.com/embedded/jetson-orin}} (64GB LPDDR5, Ampere GPU with 2048 CUDA cores and 64 Tensor Cores) configured in MaxN mode (60W), using TensorRT FP16 inference for deployment analysis. 

Segmentation quality is evaluated using standard task-specific metrics: Panoptic Quality (PQ) for panoptic segmentation, mean Average Precision (mAP) for instance segmentation, and mean Intersection-over-Union (mIoU) for semantic segmentation, with all results reported under an embedded deployment setting.


\begin{table*}[t]
\centering
\footnotesize
\setlength{\tabcolsep}{6pt}
\renewcommand{\arraystretch}{1.15}

\caption{Performance and efficiency of the proposed TokenMask head on Cityscapes for different ViT backbones.
All results are reported at $1024\times1024$ input resolution.}
\label{tab:main_city}

\begin{tabular}{l l c c c c}
\toprule
\textbf{Backbone} & \textbf{Method}
& \textbf{FPS A100} $\uparrow$
& \textbf{FPS Orin} $\uparrow$
& \textbf{GFLOPs} $\downarrow$
& \textbf{mIoU} $\uparrow$ \\
\midrule

ViT-Tiny 
& Segmenter      &  32.9  & 40.1 & 117.0 & 73.1 \\
& EoMT             & 132.0 & 42.5 & 45.6  & 69.7 \\
& \textbf{TokenMask} (Ours) & 160.0 & 53.8 & 23.2  & 68.0 \\

\midrule

ViT-Small
& Segmenter      &  15.7  & 22.1 & 285.0 & 76.6 \\
& EoMT             & 78.3  & 22.6 & 165.4 & 81.0 \\
& \textbf{TokenMask} (Ours) & 104.4 & 27.2 & 90.4  & 79.7 \\

\midrule

ViT-Base
& Segmenter   &  6.8  & 9.9  & 776.0 & 77.1 \\
& EoMT             & 37.5  & 9.6 & 628.1 & 81.8 \\
& \textbf{TokenMask} (Ours) & 51.4  & 11.9 & 356.6 & 80.4 \\

\midrule

ViT-Large
& Segmenter      &  20.8 & 3.4 & 2248.3 & 79.1 \\
& EoMT             & 17.6  & 3.6 & 1721.3 & 83.6 \\
& \textbf{TokenMask} (Ours) & 21.9  & 4.2 & 1251.6 & 82.9 \\

\bottomrule
\end{tabular}
\end{table*}

\paragraph{Training Protocol.}
All models are trained using the AdamW optimizer with identical learning rate schedules and data augmentation strategies across architectural variants. Training duration follows standard practice for each dataset: 160 epochs for ADE20K, 24 epochs for COCO-Panoptic, and 107 epochs for Cityscapes. We use $Q{=}200$ queries for ADE20K and COCO, and $Q{=}100$ queries for Cityscapes. Models are trained specifically for each target task, rather than reusing a single checkpoint across multiple objectives.
All models share identical 
initialization, decoder depth, query count, and training protocol. 

\begin{table*}[t]
\centering
\footnotesize
\setlength{\tabcolsep}{6pt}
\renewcommand{\arraystretch}{1.15}

\caption{Performance and efficiency comparison on COCO-Panoptic for instance and panoptic segmentation.
All results are reported at $640\times640$ input resolution.
GFLOPs are identical for instance and panoptic settings and therefore reported once.}
\label{tab:coco_instance_panoptic}

\begin{tabular}{l l c ccc ccc}
\toprule
\multicolumn{3}{c}{} 
& \multicolumn{3}{c}{\textbf{Instance}} 
& \multicolumn{3}{c}{\textbf{Panoptic}} \\
\cmidrule(lr){4-6} \cmidrule(lr){7-9}

\textbf{Backbone} & \textbf{Method}
& \textbf{GFLOPs} $\downarrow$
& \textbf{FPS A100} $\uparrow$
& \textbf{FPS Orin} $\uparrow$
& \textbf{mAP} $\uparrow$
& \textbf{FPS A100} $\uparrow$
& \textbf{FPS Orin} $\uparrow$
& \textbf{PQ} $\uparrow$ \\

\midrule

ViT-Tiny
& EoMT
& 18.39
& 198.5 & 148.1 & 21.5
& 202.7 & 145.9 & 32.8 \\
& \textbf{TokenMask} (Ours)
& 9.47
& 218.6 & 210.8 & 20.2
& 213.2 & 210.9 & 32.0 \\

\midrule

ViT-Small
& EoMT
& 64.14
& 136.2 & 69.3 & 29.8
& 137.4 & 66.8 & 41.9 \\
& \textbf{TokenMask} (Ours)
& 36.66
& 153.8 & 106.5 & 27.5
& 156.2 & 107.0 & 41.0 \\

\midrule

ViT-Base
& EoMT
& 237.56
& 107.2 & 34.5 & 33.8
& 107.4 & 34.8 & 46.8 \\
& \textbf{TokenMask} (Ours)
& 144.22
& 140.4 & 44.6 & 31.0
& 142.6 & 43.8 & 44.6 \\

\midrule

ViT-Large
& EoMT
& 656.43
& 52.6 & 13.3 & 40.9
& 52.6 & 12.4 & 53.3 \\
& \textbf{TokenMask} (Ours)
& 497.92
& 64.9 & 15.0 & 37.4
& 64.9 & 15.0 & 50.9 \\

\bottomrule
\end{tabular}
\end{table*}

\subsection{Main Results}
\label{sec:main}

Tables~\ref{tab:panoptic_semantic}, \ref{tab:main_city} and~\ref{tab:coco_instance_panoptic} summarize the accuracy-efficiency trade-off of TokenMask across multiple datasets and ViT backbones, and compare it to representative segmentation methods such as Mask2Former, Segmenter, and lightweight architectures like EoMT and LiPS.

Across datasets, tasks, and backbone scales, TokenMask consistently reduces computational cost while maintaining competitive segmentation accuracy. On ADE20K panoptic segmentation (Table~\ref{tab:panoptic_semantic}), this translates into 53.6

Across all datasets, tasks, and backbone scales, TokenMask systematically consistently reduces computational cost while 
and improves runtime performance while maintaining competitive segmentation accuracy. On ADE20K panoptic segmentation (Table~\ref{tab:panoptic_semantic}), compared to EoMT, this translates into 53.6\%, 48.0\%, and 44.4\% GFLOPs reductions for ViT-Tiny/Small/Base, respectively, with corresponding speedups of up to +53.7\% on Nvidia Jeston Orin. 
For ViT-Large, computation is reduced by 28.3\% with corresponding runtime gains of +15.9\% (Orin). These efficiency gains come with only moderate PQ variations (typically within 1.5-2.8 points), indicating that a large fraction of feature-space computation can be removed with limited impact on segmentation quality. Relative to Mask2Former and heavier Segmenter variants, TokenMask reduces computation by up to ~75–90\% and delivers up to ~2× inference speedup. Against lightweight architectures such as LiPS and Segmenter (Tiny/Small), TokenMask improves efficiency, reducing computation by ~20–60\% and delivering up to ~30–45\% inference speedups.


A similar trend is observed for semantic segmentation on ADE20K (Table~\ref{tab:panoptic_semantic}) compared to EoMT: on ViT-Small, GFLOPs are reduced by 47\% with a +53.3\% speedup and only a minor mIoU drop, while on ViT-Base, TokenMask reduces computation and improves runtime while matching or slightly improving mIoU.

\begin{table}[t]
\centering
\footnotesize
\setlength{\tabcolsep}{6pt}
\renewcommand{\arraystretch}{1.15}

\caption{Peak GPU memory during forward profiling (MB).
``Peak alloc.'' denotes maximum allocated CUDA memory,
and ``Peak res.'' the maximum memory reserved by the CUDA caching allocator.
Measurements use batch size 1, input resolution $640^2$,
\texttt{torch.compile=max-autotune}, and AMP FP16.}
\label{tab:peak-mem-backbones}

\begin{tabular}{l cc cc}
\toprule
&
\multicolumn{2}{c}{\textbf{EoMT}} &
\multicolumn{2}{c}{\textbf{TokenMask (Ours)}} \\
\cmidrule(lr){2-3} \cmidrule(lr){4-5}
\textbf{Backbone}
& \textbf{Peak alloc.} & \textbf{Peak res.}
& \textbf{Peak alloc.} & \textbf{Peak res.} \\
\midrule
ViT-Tiny  & 137.3  & 250.0  &  98.2  & 192.0 \\
ViT-Small & 302.2  & 488.0  & 189.6  & 228.0 \\
ViT-Base  & 763.6  & 1168.0 & 499.2  & 682.0 \\
ViT-Large & 1746.6 & 2210.0 & 1363.0 & 1446.0 \\
ViT-Giant & 5277.0 & 5908.0 & 4667.8 & 4820.0 \\
\bottomrule
\end{tabular}
\end{table}

These computational reductions translate directly into lower activation memory. As reported in Table~\ref{tab:peak-mem-backbones}, peak allocated memory decreases consistently across scales (e.g., 302.2$\rightarrow$189.6 MB for ViT-Small and 763.6$\rightarrow$499.2 MB for ViT-Base). For larger backbones, the absolute savings become substantial (e.g., over 380 MB reduction for ViT-Large).

On COCO-Panoptic (Table~\ref{tab:coco_instance_panoptic}), TokenMask exhibits similar trends for both instance and panoptic tasks. For ViT-Tiny and ViT-Small, compute is nearly halved (18.39$\rightarrow$9.47 and 64.14$\rightarrow$36.66 GFLOPs) and throughput increases consistently. For instance segmentation, inference speed on Orin increases from 148.1 to 210.8 FPS (+42.3\%) for ViT-Tiny and from 69.3 to 106.5 FPS (+53.7\%) for ViT-Small. 
For panoptic segmentation, Orin FPS improves from 145.9$\rightarrow$210.9 (+44.5\%) and 66.8$\rightarrow$107.0 (+60.2\%). Accuracy remains close to the baseline across scales (e.g., PQ 32.8$\rightarrow$32.0 and 41.9$\rightarrow$41.0), indicating that the reformulation primarily restructures computation rather than substantially altering recognition capacity.

Under higher-resolution inputs, Cityscapes confirms the same scaling behavior (Table~\ref{tab:main_city}). Compared to EoMT, for ViT-Base, GFLOPs decrease from 628.06 to 356.62 (-43.2\%), with FPS improving from 9.6$\rightarrow$11.9 on Orin (+24.0\%), while mIoU remains competitive (81.8$\rightarrow$80.4). Similar gains are observed for ViT-Small (22.6$\rightarrow$27.2 FPS on Orin) with limited accuracy change (81.0$\rightarrow$79.7). In contrast to Segmenter, TokenMask reduces computation by more than 2× across backbones (e.g., 776.0$\rightarrow$356.6 GFLOPs on ViT-Base), while improving FPS and maintaining comparable or better mIoU.

\begin{figure}[t]
    \centering
    \includegraphics[width=\linewidth]{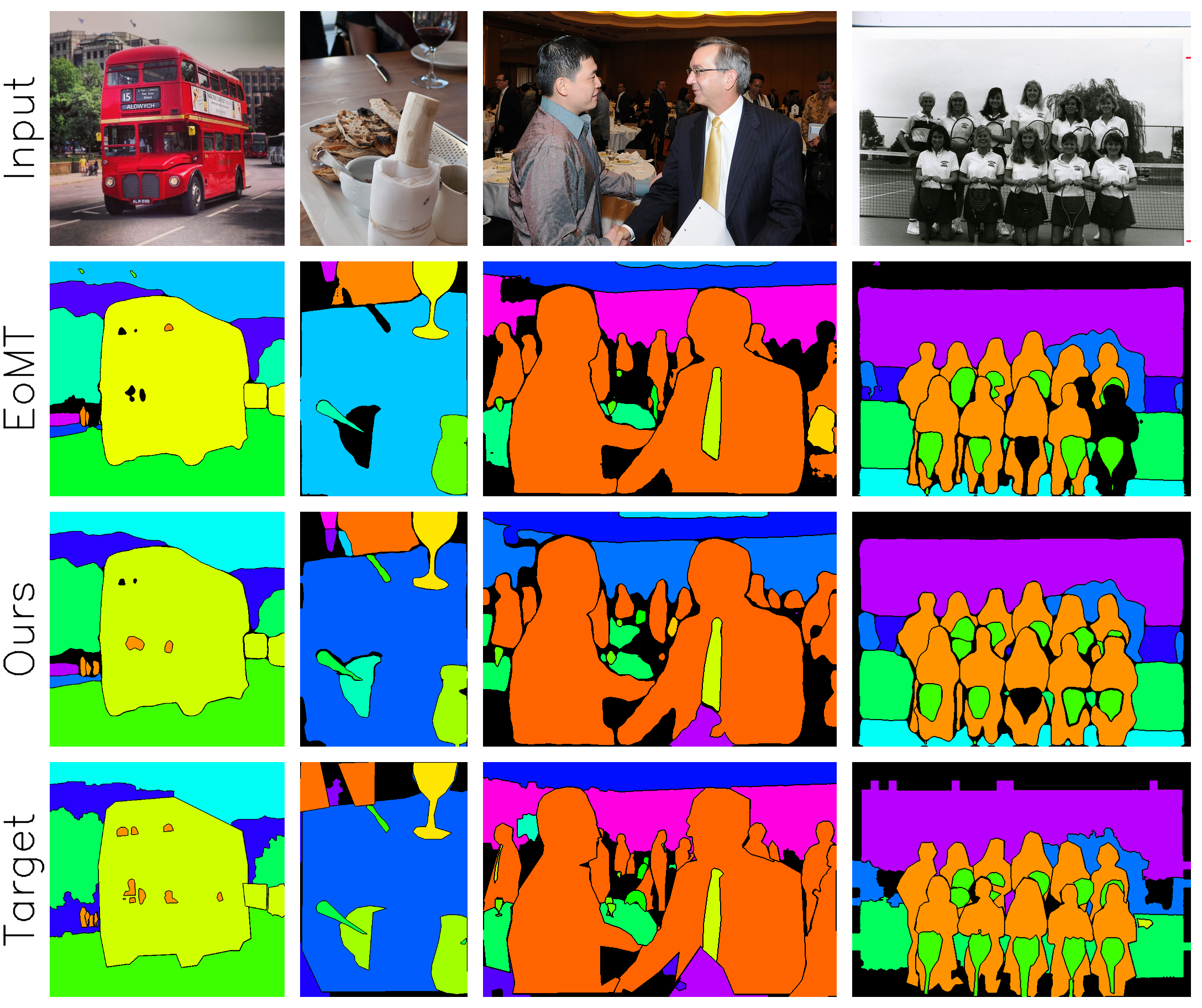}
    \caption{Qualitative comparison on panoptic segmentation.}
    \label{fig:qualitative}
\end{figure}

Qualitative comparisons in Figure~\ref{fig:qualitative} further support these findings: across diverse scene layouts and object scales, TokenMask produces segmentation results that are visually consistent with the ground truth. 
Both small instances and large structural regions are accurately delineated, indicating that the reduced computation does not compromise spatial coherence or instance separation.

%% file: content/5_ablations.tex
\subsection{Ablation Studies}
\label{subsectuib:ablation}

We conduct a structured ablation study to isolate the impact of individual design choices within the mask prediction pipeline. 
Specifically, we study: 
\textbf{(1)} relocating interpolation from feature space to logit space;
\textbf{(2)} removing upsampling entirely; 
\textbf{(3)} varying output stride as a deployment-level parameter; and
\textbf{(4)} numerical consistency under mixed-precision TensorRT inference. Unless stated otherwise, all ablations use ADE20K (panoptic), 640$\times$640 inputs, on NVIDIA A100 GPUs.

\paragraph{Relocating Interpolation: Feature vs Logit Space.}

We first retain the classical feature-map mask head and modify only the location of spatial interpolation. In the baseline EoMT formulation, interpolation is applied to $C$-channel feature maps prior to mask scoring. We instead apply interpolation to the $Q$-channel mask logits after scoring, while keeping all other components unchanged. As shown in Table~\ref{tab:eomt_upsample_ablation_multi}, switching from feature-space to logit-space upsampling leads to a consistent but moderate PQ decrease (between $-1.1$ and $-2.2$ points) across all backbones. However, this accuracy drop remains limited compared to the substantial reductions in computational cost and the improvement in inference speed. This corresponds to FLOPs reductions of 53.6\%, 48.0\%, 44.4\%, and 28.3\% for ViT-Tiny, ViT-Small, ViT-Base, and ViT-Large, respectively. 
These results indicate that most of the spatial refinement provided by feature-space interpolation is redundant.

\begin{table}[t]
\centering
\footnotesize
\setlength{\tabcolsep}{4pt}
\renewcommand{\arraystretch}{1.15}

\caption{Ablation of upsampling strategies applied to the EoMT. 
}
\label{tab:eomt_upsample_ablation_multi}

\begin{tabular}{l l l c c c}
\toprule
\textbf{Method} & \textbf{Backbone} & \textbf{Upsampling} & \textbf{FPS} $\uparrow$ & \textbf{GFLOPs} $\downarrow$ & \textbf{PQ} $\uparrow$ \\
\midrule

\multirow{12}{*}{
\parbox[c]{1.1cm}{\centering \rotatebox{90}{EoMT}}
}

& \multirow{3}{*}{ViT-Tiny} 
& Feature & 189.0 & 20.85 & 29.5 \\
& & Logit   & 215.7 & 9.68 & 28.4 \\
& & None    & 217.0 & 9.58 & 28.1 \\
\cmidrule(lr){2-6}

& \multirow{3}{*}{ViT-Small} 
& Feature & 129.3 & 71.69 & 38.0 \\
& & Logit   & 153.7 & 37.28 & 35.8 \\
& & None    & 154.5 & 37.18 & 35.7 \\
\cmidrule(lr){2-6}

& \multirow{3}{*}{ViT-Base} 
& Feature & 94.8 & 263.15 & 40.5 \\
& & Logit   & 135.3 & 146.33 & 39.0 \\
& & None    & 135.6 & 146.23 & 38.9 \\
\cmidrule(lr){2-6}

& \multirow{3}{*}{ViT-Large} 
& Feature & 48.5 & 699.89 & 46.6 \\
& & Logit   & 61.4 & 501.51 & 45.4 \\
& & None    & 62.3 & 501.41 & 45.0 \\

\bottomrule
\end{tabular}
\end{table}

\begin{table}[t]
\centering
\footnotesize
\setlength{\tabcolsep}{4pt}
\renewcommand{\arraystretch}{1.15}

\caption{Ablation study on the role of logit-space upsampling in the proposed TokenMask head. 
}
\label{tab:ours_token_multi}

\begin{tabular}{l l l c c c}
\toprule
\textbf{Method} 
& \textbf{Backbone} 
& \textbf{Upsampling} 
& \textbf{FPS} $\uparrow$ 
& \textbf{GFLOPs} $\downarrow$ 
& \textbf{PQ} $\uparrow$ \\
\midrule

\multirow{8}{*}{
\parbox[c]{1.1cm}{\centering \rotatebox{90}{\textbf{TokenMask} (Ours)}}
}

& \multirow{2}{*}{ViT-Tiny} 
& Logit & 202.9 & 9.68 & 27.7 \\
& & None  & 212.8 & 9.58 & 28.1 \\
\cmidrule(lr){2-6}

& \multirow{2}{*}{ViT-Small} 
& Logit & 150.8 & 37.28 & 35.2 \\
& & None  & 154.3 & 37.18 & 35.5 \\
\cmidrule(lr){2-6}

& \multirow{2}{*}{ViT-Base} 
& Logit & 131.6 & 146.33 & 39.0 \\
& & None  & 136.4 & 146.23 & 38.0 \\
\cmidrule(lr){2-6}

& \multirow{2}{*}{ViT-Large} 
& Logit & 61.9 & 501.51 & 45.1 \\
& & None  & 61.9 & 501.41 & 45.1 \\

\bottomrule
\end{tabular}
\end{table}

\paragraph{Removing Upsampling Entirely.}

We next evaluate the extreme configuration where no spatial upsampling is applied and masks are produced directly at patch-grid resolution. This removes both feature-space and logit-space interpolation. The purpose of this experiment is to assess whether explicit spatial refinement is fundamentally required for competitive segmentation performance. Results in Tables~\ref{tab:eomt_upsample_ablation_multi} and~\ref{tab:ours_token_multi} show 
that completely removing the upsampling operation yields only marginal additional computational savings compared to logit-space interpolation (around 0.1 GFLOPs across all backbones), corresponding to less than 1\% relative reduction. This indicates that most of the computational overhead is already eliminated when shifting interpolation to logit space. We also observe a small but consistent PQ decrease (between $-0.1$ and $-0.4$ points), which, combined with the negligible additional computational savings, suggests that logit-space interpolation is nearly cost-free yet still beneficial for maintaining segmentation quality.





\begin{table}[t]
\centering
\footnotesize
\setlength{\tabcolsep}{6pt}
\renewcommand{\arraystretch}{1.15}

\caption{Effect of output stride for TokenMask with a ViT-Small.}
\label{tab:stride_study}

\begin{tabular}{c c c c}
\toprule
\textbf{Output Stride} 
& \textbf{FPS} $\uparrow$ 
& \textbf{GFLOPs} $\downarrow$ 
& \textbf{PQ} $\uparrow$ \\
\midrule
1  & 154.0 & 38.73 & 33.9 \\
4  & 196.0 & 37.19 & 35.1 \\
8  & 199.7 & 37.12 & 33.9 \\
16 & 197.8 & 37.09 & 34.5 \\
\bottomrule
\end{tabular}
\end{table}

\paragraph{Output Stride as a Deployment Knob.}

We further analyze the effect of varying the mask output stride when using logit-space upsampling. Instead of fixing the output resolution, we evaluate multiple stride values to study the trade-off between spatial resolution and computational cost. As shown in Table~\ref{tab:stride_study}, stride $4$ emerges as the best trade-off, achieving the highest PQ (35.1) while substantially improving inference speed compared to stride $1$ (+27\% FPS) and reducing computational cost. Although larger strides (8 and 16) provide marginal additional FLOPs reductions (below 0.3\%) and slight throughput gains, they consistently degrade PQ (up to $-1.2$ points).

%% file: content/6_conclusions.tex
\section{Conclusions}

We introduced TokenMask, a token-space mask head that reformulates query-based segmentation by computing mask logits directly from query-token affinities, thereby eliminating explicit image-space feature reconstruction. 
By aligning mask prediction with the intrinsic token representation of ViTs and leveraging GEMM-friendly operations, TokenMask provides a hardware-conscious and structurally streamlined approach to semantic, instance, and panoptic segmentation. 
Our analysis demonstrated that relocating interpolation from feature space to logit space substantially improves efficiency, and that even removing upsampling entirely remains viable in many regimes. These findings clarify the role of upsampling in query-based segmentation and expose favorable scaling behavior with respect to embedding dimension and spatial resolution.
 
Across all evaluated datasets and over multiple ViT backbone scales, TokenMask consistently reduces computational complexity, increases throughput, and lowers peak memory usage while maintaining competitive segmentation accuracy. Compared to EoMT, TokenMask reduces computational cost by up to 53.6\%, improves runtime by up to 1.5$\times$, and decreases peak memory usage by several hundreds of megabytes across backbone scales, while incurring only minor accuracy variations (typically within 1.5-2.8 PQ points). 

Overall, our results suggest that efficient segmentation with ViTs does not require explicit spatial reconstruction, opening new perspectives for lightweight, deployment-oriented design in embedded vision systems.

%% file: main.bib
@String(CVPR= {IEEE Conf. Comput. Vis. Pattern Recog.})

@String(AAAI = {AAAI})

@String(CVPR  = {CVPR})

@article{galagain2025semanticslamreadyembedded,
  title={Is Semantic SLAM Ready for Embedded Systems? A Comparative Survey},
  author={Galagain, Calvin and Poreba, Martyna and Goulette, Fran{\c{c}}ois},
  journal={arXiv preprint arXiv:2505.12384},
  year={2025}
}

@article{dosovitskiy2020image,
  title={An image is worth 16x16 words: Transformers for image recognition at scale},
  author={Dosovitskiy, Alexey and Beyer, Lucas and Kolesnikov, Alexander and Weissenborn, Dirk and Zhai, Xiaohua and Unterthiner, Thomas and Dehghani, Mostafa and Minderer, Matthias and Heigold, Georg and Gelly, Sylvain and others},
  journal={Int. Conf. Learn. Represent.},
  year={2021}
}

@inproceedings{Carion2020EndtoEndOD,
  title={End-to-end object detection with transformers},
  author={Carion, Nicolas and Massa, Francisco and Synnaeve, Gabriel and Usunier, Nicolas and Kirillov, Alexander and Zagoruyko, Sergey},
  booktitle={Proc. of the Eur. Conf. on Computer Vision},
  pages={213-229},
  year={2020},
  organization={}
}

@article{Cheng2021PerPixelCI,
  title={Per-pixel classification is not all you need for semantic segmentation},
  author={Cheng, Bowen and Schwing, Alex and Kirillov, Alexander},
  journal={Advances in Neural Information Processing Systems},
  volume={34},
  pages={17864-17875},
  year={2021}
}

@inproceedings{Cheng2021MaskedattentionMT,
  title={Masked-attention mask transformer for universal image segmentation},
  author={Cheng, Bowen and Misra, Ishan and Schwing, Alexander G and Kirillov, Alexander and Girdhar, Rohit},
  booktitle={Proc. of the IEEE/CVF Conf. on Computer Vision and Pattern Recognition},
  pages={1290-1299},
  year={2022}
}

@inproceedings{Jain2022OneFormerOT,
  title={Oneformer: One transformer to rule universal image segmentation},
  author={Jain, Jitesh and Li, Jiachen and Chiu, Mang Tik and Hassani, Ali and Orlov, Nikita and Shi, Humphrey},
  booktitle={Proc. of the IEEE/CVF Conf. on Computer Vision and Pattern Recognition},
  pages={2989-2998},
  year={2023}
}

@inproceedings{Kerssies2025YourVI,
  title={Your vit is secretly an image segmentation model},
  author={Kerssies, Tommie and Cavagnero, Niccolo and Hermans, Alexander and Norouzi, Narges and Averta, Giuseppe and Leibe, Bastian and Dubbelman, Gijs and de Geus, Daan},
  booktitle={Proc. of the Computer Vision and Pattern Recognition},
  pages={25303-25313},
  year={2025}
}

@article{Cordts2016TheCD,
  title={The Cityscapes Dataset for Semantic Urban Scene Understanding},
  author={Marius Cordts and Mohamed Omran and Sebastian Ramos and Timo Rehfeld and Markus Enzweiler and Rodrigo Benenson and Uwe Franke and Stefan Roth and Bernt Schiele},
  journal={IEEE Conf. on Computer Vision and Pattern Recognition },
  year={2016},
  pages={3213-3223},
  url={https://api.semanticscholar.org/CorpusID:502946}
}

@inproceedings{zhou2017scene,
  title={Scene parsing through ade20k dataset},
  author={Zhou, Bolei and Zhao, Hang and Puig, Xavier and Fidler, Sanja and Barriuso, Adela and Torralba, Antonio},
  booktitle={Proc. of the IEEE Conf. on Computer Vision and Pattern Recognition},
  pages={633-641},
  year={2017}
}

@inproceedings{lin2014microsoft,
  title={Microsoft coco: Common objects in context},
  author={Lin, Tsung-Yi and Maire, Michael and Belongie, Serge and Hays, James and Perona, Pietro and Ramanan, Deva and Doll{\'a}r, Piotr and Zitnick, C Lawrence},
  booktitle={Eur. Conf. on Computer Vision},
  pages={740-755},
  year={2014},
  organization={Springer}
}

@inproceedings{SETR,
    title={Rethinking Semantic Segmentation from a Sequence-to-Sequence Perspective with Transformers}, 
    author={Zheng, Sixiao and Lu, Jiachen and Zhao, Hengshuang and Zhu, Xiatian and Luo, Zekun and Wang, Yabiao and Fu, Yanwei and Feng, Jianfeng and Xiang, Tao and Torr, Philip H.S. and Zhang, Li},
    booktitle={CVPR},
    year={2021}
}

@article{Zhang_2024,
   title={Vision Transformers: From Semantic Segmentation to Dense Prediction},
   volume={132},
   ISSN={1573-1405},
   url={http://dx.doi.org/10.1007/s11263-024-02173-w},
   DOI={10.1007/s11263-024-02173-w},
   number={12},
   journal={International Journal of Computer Vision},
   publisher={Springer Science and Business Media LLC},
   author={Zhang, Li and Lu, Jiachen and Zheng, Sixiao and Zhao, Xinxuan and Zhu, Xiatian and Fu, Yanwei and Xiang, Tao and Feng, Jianfeng and Torr, Philip H. S.},
   year={2024},
   month=jul, pages={6142–6162} }

@misc{xie2021segformersimpleefficientdesign,
      title={SegFormer: Simple and Efficient Design for Semantic Segmentation with Transformers}, 
      author={Enze Xie and Wenhai Wang and Zhiding Yu and Anima Anandkumar and Jose M. Alvarez and Ping Luo},
      year={2021},
      eprint={2105.15203},
      archivePrefix={arXiv},
      primaryClass={cs.CV},
      url={https://arxiv.org/abs/2105.15203}, 
}

@inproceedings{strudel2021segmenter,
  title={Segmenter: Transformer for semantic segmentation},
  author={Strudel, Robin and Garcia, Ricardo and Laptev, Ivan and Schmid, Cordelia},
  booktitle={Proceedings of the IEEE/CVF international conference on computer vision},
  pages={7262--7272},
  year={2021}
}

@ARTICLE{StructToken,
  author={Lin, Fangjian and Liang, Zhanhao and Wu, Sitong and He, Junjun and Chen, Kai and Tian, Shengwei},
  journal={IEEE Transactions on Circuits and Systems for Video Technology}, 
  title={StructToken: Rethinking Semantic Segmentation With Structural Prior}, 
  year={2023},
  volume={33},
  number={10},
  pages={5655-5663},
  doi={10.1109/TCSVT.2023.3252807}}

@misc{zhang2022segvitsemanticsegmentationplain,
      title={SegViT: Semantic Segmentation with Plain Vision Transformers}, 
      author={Bowen Zhang and Zhi Tian and Quan Tang and Xiangxiang Chu and Xiaolin Wei and Chunhua Shen and Yifan Liu},
      year={2022},
      eprint={2210.05844},
      archivePrefix={arXiv},
      primaryClass={cs.CV},
      url={https://arxiv.org/abs/2210.05844}, 
}

@misc{zhang2023segvitv2exploringefficientcontinual,
      title={SegViTv2: Exploring Efficient and Continual Semantic Segmentation with Plain Vision Transformers}, 
      author={Bowen Zhang and Liyang Liu and Minh Hieu Phan and Zhi Tian and Chunhua Shen and Yifan Liu},
      year={2023},
      eprint={2306.06289},
      archivePrefix={arXiv},
      primaryClass={cs.CV},
      url={https://arxiv.org/abs/2306.06289}, 
}

@inproceedings{Dong2023HeadFreeLS,
  title={Head-free lightweight semantic segmentation with linear transformer},
  author={Dong, Bo and Wang, Pichao and Wang, Fan},
  booktitle={Proc. of the AAAI conf. on artificial intelligence},
  volume={37},
  number={1},
  pages={516--524},
  year={2023}
}

@misc{galagain2026lipslightweightpanopticsegmentation,
      title={LiPS: Lightweight Panoptic Segmentation for Resource-Constrained Robotics}, 
      author={Calvin Galagain and Martyna Poreba and François Goulette and Cyrill Stachniss},
      year={2026},
      eprint={2604.00634},
      archivePrefix={arXiv},
      primaryClass={cs.RO},
      url={https://arxiv.org/abs/2604.00634}, 
}
